\documentclass{svproc}
\usepackage{epsfig}
\graphicspath{ {./Pic/} }

\usepackage{multirow}

\usepackage{url}

\begin{document}
\mainmatter              % start of a contribution
\title{Simultaneous prediction of hand gestures, handedness, and hand keypoints using thermal images}
\titlerunning{Hand gesture, handedness, and hand keypoints detection}  % abbreviated title (for running head)
\author{Sichao Li\inst{1} \and Sean Banerjee\inst{1}
Natasha Kholgade Banerjee\inst{1} \and Soumyabrata Dey\inst{1}}
\authorrunning{Sichao Li et al.} % abbreviated author list (for running head)
%Craig Chambers, Kim B. Bruce, and Elisa Bertino}
%
\institute{Clarkson University, 8 Clarkson Avenue NY 13699, USA,\\
\email{sdey@clarkson.edu},\\ WWW home page:
\texttt{https://www.clarkson.edu/people/soumyabrata-dey}}
%\and
%Universit\'{e} de Paris-Sud,
%Laboratoire d'Analyse Num\'{e}rique, B\^{a}timent 425,\\
%F-91405 Orsay Cedex, France}

\maketitle              % typeset the title of the contribution

\begin{abstract}
Hand gesture detection is a well-explored area in computer vision with applications in various forms of Human-Computer Interactions. In this work, we propose a technique for simultaneous hand gesture classification, handedness detection, and hand keypoints localization using thermal data captured by an infrared camera. Our method uses a novel deep multi-task learning architecture that includes shared encoder-decoder layers followed by three branches dedicated for each mentioned task. We performed extensive experimental validation of our model on an in-house dataset consisting of 24 users' data. The results confirm higher than $98\%$ accuracy for gesture classification, handedness detection, and fingertips localization, and more than $91\%$ accuracy for wrist points localization.
% We would like to encourage you to list your keywords within
% the abstract section using the \keywords{...} command.
\end{abstract}

\section{Introduction}
\label{sec:intro}

With the fast-changing technology landscape, the use of interconnected smart devices equipped with sensors has become progressively popular. Smart devices have use in various important applications such as smart-home, self-driving cars, smart-infrastructure, and smart cities. However, the interactions with the smart devices are still not easy because users often need to learn and remember different settings and operation manuals specific to each device. As we will be more dependent on technology in the near future, the method of interactions with the smart devices needs to be more user friendly.

Because of recent technological developments, gesture-activated and voice-command-based devices are becoming available in the market. These devices provide the users with natural ways of interacting with them reducing the trouble of remembering complex setting information. Hand-gesture-based Human-Computer Interaction (HCI) is one of the major fields in computer vision that has been studied for many years. However, most of the works explored hand detection \cite{yolse,trace_rgb1,Park2012HandDA,Xu2020AccurateHD,Li2013PixelLevelHD,Gao2020RobustRH} and gesture identification \cite{kmeans_ml,gesture_rgb1,gesture_rgb_skin,real_time} tasks use data from RGB (red, green, and blue) cameras.  Many of these works use a skin color database to segment the hand regions from the rest of the scene and traditional machine learning or deep learning techniques for gesture classification \cite{kmeans_ml,old_dl,gesture_rgb1,Gao2020RobustRH}.  

Recently, other sensor modalities such as depth and thermal cameras are becoming widely available. Regardless, there have been a comparatively limited number of attempts for hand-related applications such as hand gesture classification using depth \cite{fast_depth,thermal_depth,palm_depth} and thermal data \cite{ahd,Tracking_thermal,thermal1,thermal_data,Ballow2022,Gately2020}. While RGB-data-based methods can suffer from problems such as lighting condition variations and skin color variations that can negatively impact the accuracy of the hand detection method, the thermal-data-based approaches are less affected by those variations and can complement RGB-based techniques. Therefore, extensive study for hand gesture detection using thermal data is necessary to understand the capability of a complementary data modality and for a possible robust future approach combining both color and thermal data modalities.

In this paper, we propose a novel deep learning architecture for simultaneous gesture classification, handedness detection, and hand keypoints localization using thermal images. The deep learning (DL) model uses a shared encoder-decoder component followed by three branches for gesture classification, handedness detection, and fingertips \& wrist points localization. The network is trained by backpropagating the error estimated using a joint loss function. Furthermore, we introduced an intelligent post-processing step that utilizes the insights from the other two branches of the DL network for refining the hand keypoints localization results. We summarize the contributions of our work below.

%\begin{enumerate}
  \textbf{1.} We prepare a new dataset consisting of thermal imaging data from 24 users and 10 gestures per user.
  
  \textbf{2.} We propose a novel multi-task DL network architecture that performs gesture classification, left- and right-hand detection, and hand keypoints localization. To the best of our knowledge, this is the first work that can perform all three tasks using a single network.    
  
  \textbf{3.} We demonstrate through experimental validation the superior performance of our model on all three tasks. On average, the accuracy of the model is $>98\%$ for gesture classification, handedness detection, and fingertips localization. The wrist points localization accuracy is $> 91\%$. 
  
  \textbf{4.} Instead of a threshold-based finger point localization, we introduce an adaptive filtering technique that utilizes the insight learned in the other branches of the network. 
%\end{enumerate}

\begin{figure*}[t]
  \centering
  \centerline{\epsfig{figure=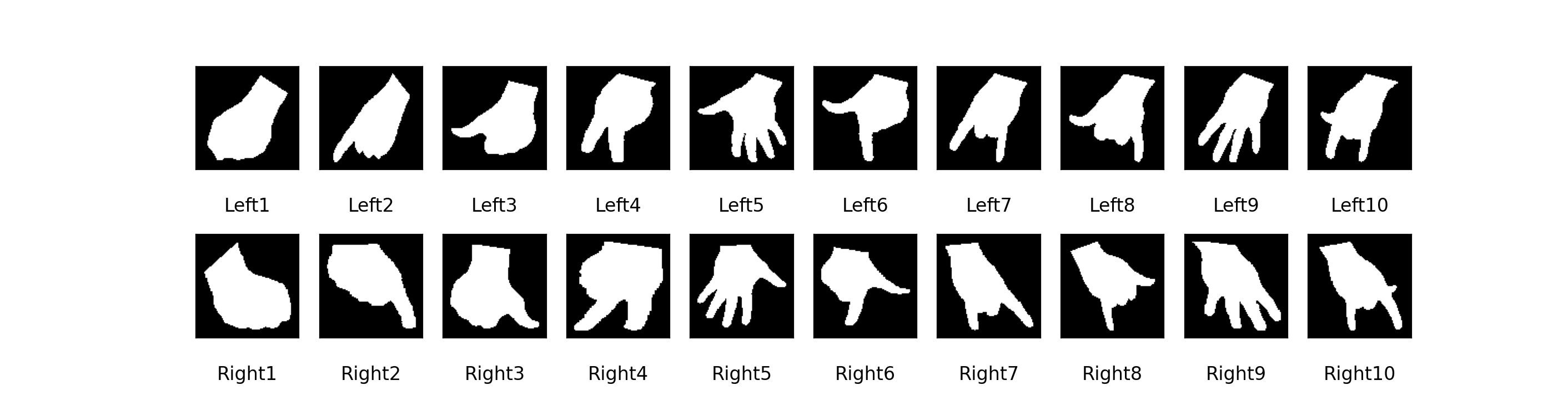,width=15cm}}
  \caption{10 different gestures for left (top row) and right (bottom row) hands.}
  \label{figure:all gestures}
\end{figure*}

\begin{table*}[t]
\begin{center}
\caption{Dataset sample counts per gesture($G_{\#}$) and left/right hands.} \label{table:dataset summary}
\begin{tabular}{|c|c|c|c|c|c|c|c|c|c|c|c|c|}
  \hline
    & G1 & G2 & G3 & G4 & G5 & G6 & G7 & G8 & G9 & G10 & Left & Right\\
  \hline
  Train & 4780 & 4700 & 4630 & 4520 & 5510 & 5390 & 4620 & 4570 & 4668 & 4938 & 23636 & 24690\\
  Test & 970 & 1120 & 1120 & 1110 & 1120 & 1120 & 1090 & 1100 & 1120 & 1100 & 5570 & 5400\\
  Total & 5750 & 5820 & 5750 & 5630 & 6630 & 6510 & 5710 & 5670 & 5788 & 6038 & 29206 & 30090\\
  \hline
%\label{table:dataset summary}  
\end{tabular}
\end{center}
\end{table*}

\begin{figure}[t]

\begin{minipage}[b]{.42\linewidth}
  \centering
  \centerline{\epsfig{figure=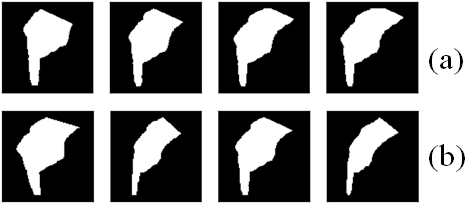,width=3.8cm}}
  %\centerline{(a) Forearm length}\medskip
\end{minipage}
\begin{minipage}[b]{.52\linewidth}
  \centering
  \centerline{\epsfig{figure=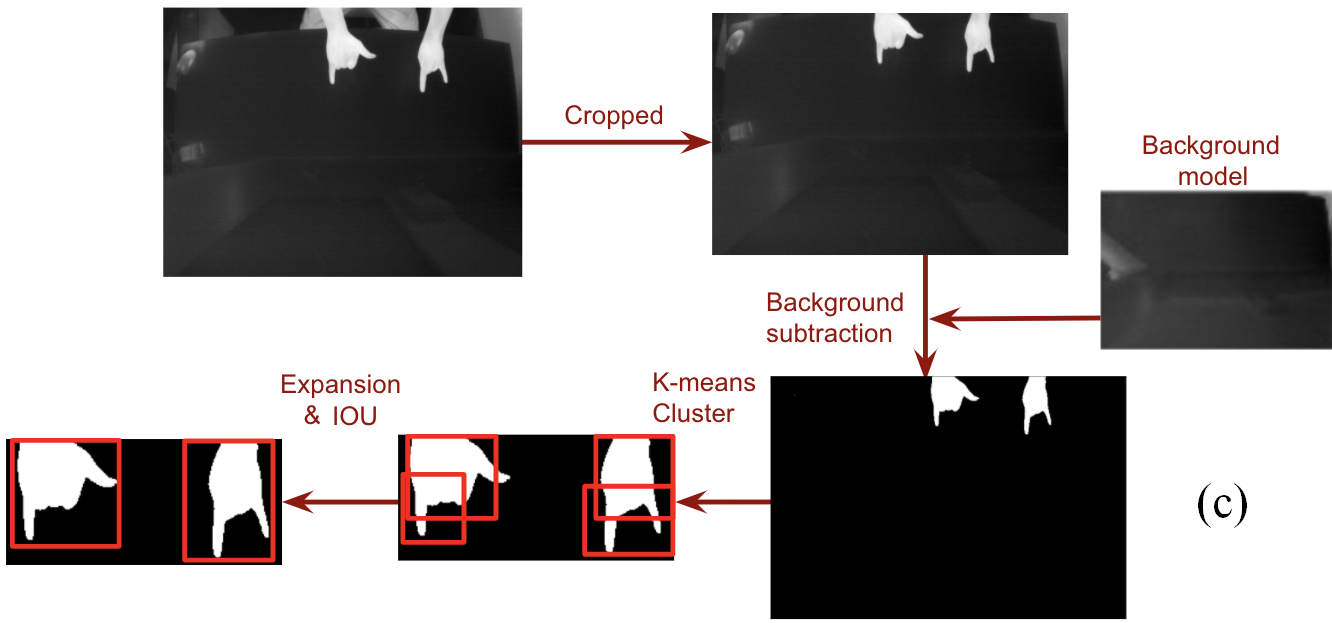,width=7.2cm}}
  %\centerline{(b) Rotation and crop}\medskip
\end{minipage}
\hfill
\caption{ Example of data augmentation. (a) Forearm cropping, (b) rotation. (c) Automatic hand-segmentation during testing.}
\label{figure:dataPreprocessing}
\end{figure}

%\begin{figure}[t]
%  \centering
%  \centerline{\epsfig{figure=data_pre.png,width=8.5cm}}
%  \caption{Test data preparation framework.}
%  \label{figure:hand detection}
%\end{figure}

\begin{figure*}[t]
  \centering
  \centerline{\epsfig{figure=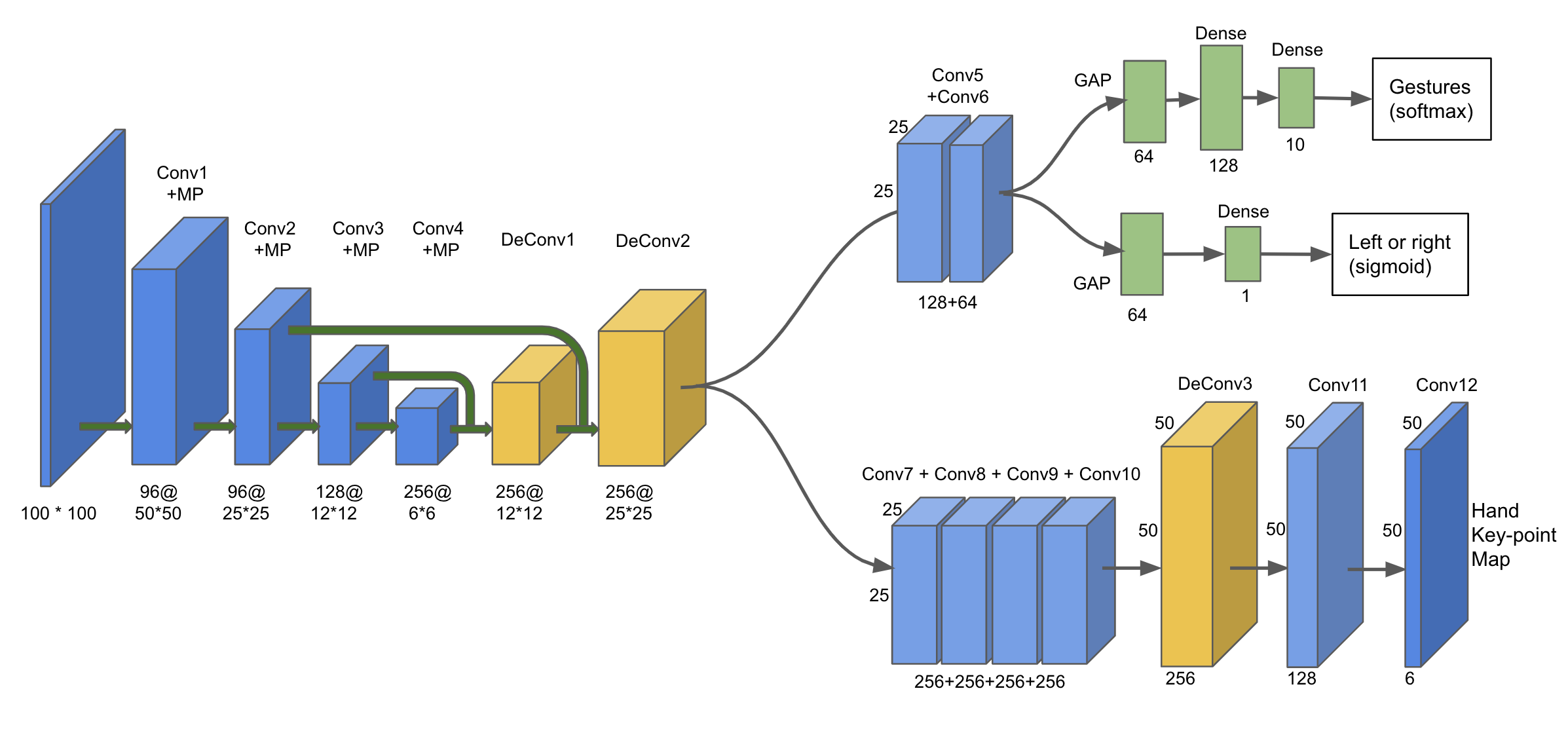, width=12cm}}
  \caption{The architecture used for gesture classification, left-right-hand detection, and hand keypoints localization.}
  \label{figure:DNN model}
\end{figure*}

\section{DATASET}
We collected a customized thermal imaging dataset to train and validate our model on all three proposed tasks. The dataset details are provided below.

\textbf{Data collection:} Data is collected from 24 users with a Sierra Olympic Viento-G thermal camera. Each user is recorded at 30fps (frames per second) performing 10 different gestures using left and right hands. All the data is captured in indoor conditions (temperature between $65^{\circ}$ to $70^{\circ}$ F) and stored as 16 bit 640×480 TIFF image sequences. The camera is fixed to a wooden stand and oriented downwards focusing on a table. Users are required to turn their palms towards the tabletop and the back of their hands towards the camera. In total, we collected 59296 frames; 29206 left-hand frames, and 30090 right-hand frames. Table \ref{table:dataset summary} summarize the dataset information and figure \ref{figure:all gestures} illustrates all gestures used in this study.    

\textbf{Data preparation:} Given the recorded frames, we go through a sequence of steps to prepare the data for our experiments. First, we crop the images to a predefined $640 \times 440$ pixel region so that only the tabletop is visible. Next, the training data is prepared by running a python script that allows manual selection of the fingertips and wrist points, cropping of the hand region, and segmentation of the images into binary foreground-background regions (background pixels = 0 and foreground pixels = 1). Finally, two data augmentation techniques, such as rotation and variable-length forearm inclusion, are applied to each image. The model is trained with a variable-length forearm because in the test scenario a user can wear clothing of different sleeve lengths occluding different lengths of the forearm. The model needs to learn to ignore this variation and identify the gesture correctly irrespective of the length of the forearm. We used 10  different lengths of the forearm per image. Figure \ref{figure:dataPreprocessing}-(a-b) illustrates the data augmentations.

During the test scenario, the hand images are generated by an automatic algorithm inspired by \cite{Ballow2022}. The algorithm use background subtraction to detect binary hand regions, k-means clustering to isolate each hand, crop the image to tightly include each hand region, and resize it to a 100x100 pixel image. Figure \ref{figure:dataPreprocessing}-(c) shows all the steps.

\section{Methods}
We propose a novel DL network architecture for learning the three proposed tasks through a joint loss function. The fingertips are further refined using the detection results of the gesture classification and handedness detection branches. The whole process is described below in detail.   

\subsection{Model structure}
The architecture of the proposed model is presented in Figure \ref{figure:DNN model}. The model expects a fixed size input of size $100 \times 100$. The shared part of the model is similar to a U-Net \cite{u-net} architecture, and it consists of four convolutional layers, four max-pooling layers, and two up-convolutional layers. The encoder part of the network uses a series of max-pooling and convolutional layers to down-sample the resolution to $6 \times 6$. This is followed by two up-convolution layers with skip connections to increase the resolution back to $25 \times 25$. 

The first two branches of the network perform the gesture classification and left-right hand detection tasks. They share two convolutional layers before separating into different paths. The gesture classification path uses a global average pooling layer followed by two dense layers. The handedness detection path uses a global average pooling layer and a dense output layer. 

The last branch is responsible for the fingertips and wrist points localization task. This consists of four convolutional layers followed by an up-convolution and two convolutional layers. The final output dimensions of this block are $50 \times 50 \times 6$. The first five channels predict the fingertip locations, and the sixth channel predicts the wrist points. This is described in the subsection \ref{subsec:hand keypoints}.   

All convolutional layers use kernel size $3 \times 3$, stride 1, and padding `same'. All up-convolution layers use kernel size $3 \times 3$ and stride 2. Our model uses `ReLU' activation function. We use stochastic gradient descent (SGD) algorithm with a learning rate of 0.001, weight decay $1e^{-3}$, and momentum of 0.95. A batch normalization \cite{batchnormal} is used between each convolution layer and activation layer.

%{\color{red} "We need batch normalization and other details here"}

\subsection{Hand keypoint detection}
\label{subsec:hand keypoints}
\begin{figure}[t]
  \centering
  \centerline{\epsfig{figure=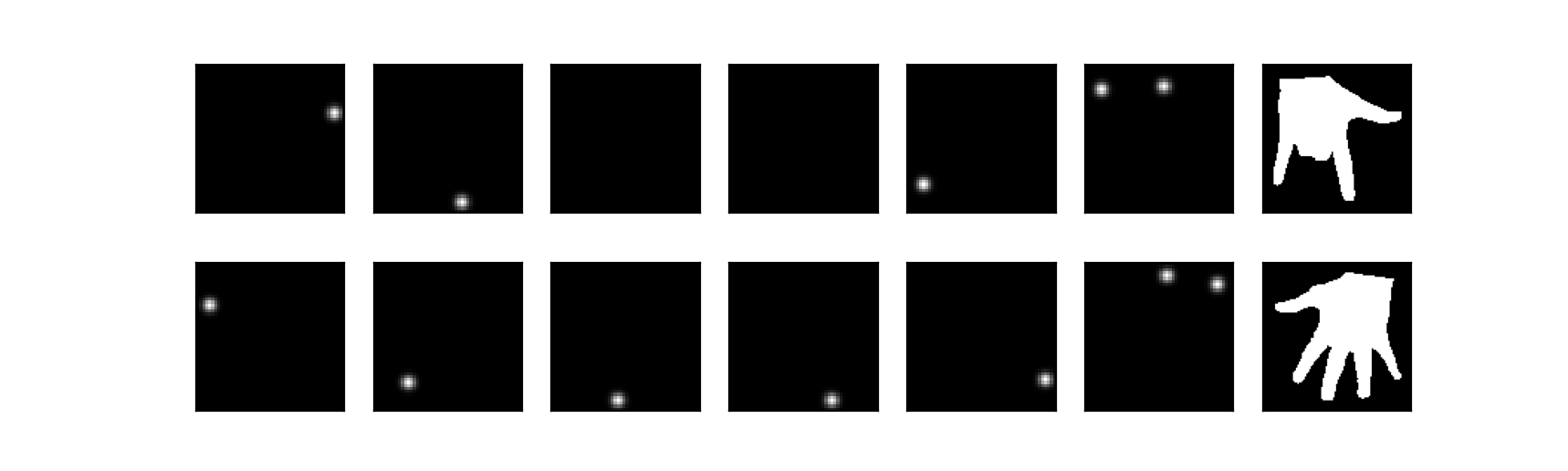,width=8.5cm}}
  \caption{Examples ground-truth output maps for right-hand gesture 10 (top row) and left-hand gesture 5 (bottom row).}
  \label{figure:hand keypoints}
\end{figure}

We trained the network to predict the hand keypoints in a $50 \times 50 \times 6$ output map. The first five channels in the map are trained to localize five fingertips in the sequence of thumb, index, middle, ring, and little finger. The last channel is dedicated to the two wrist points. Ground-truth output maps are created with the following rules. Depending on the gesture, a few fingers will be visible while others will be occluded in an input image. For an output channel, if the corresponding finger is not visible, we set all pixels to $0$s. Otherwise, a two-dimensional Gaussian with variance $1.5$ is used to set the pixel values at the fingertip location and around. Similarly, in the 6th channel, the pixel values at and around the two wrist point locations are assigned using the same Gaussian distribution. Two sets of example ground-truth maps are shown in Figure \ref{figure:hand keypoints} for better understanding. The ground-truth maps are compared against predicted output maps to estimate loss. A similar approach is used in \cite{yolse} for fingertips detection from color images.   

\begin{figure}[t]
  \centering
  \centerline{\epsfig{figure=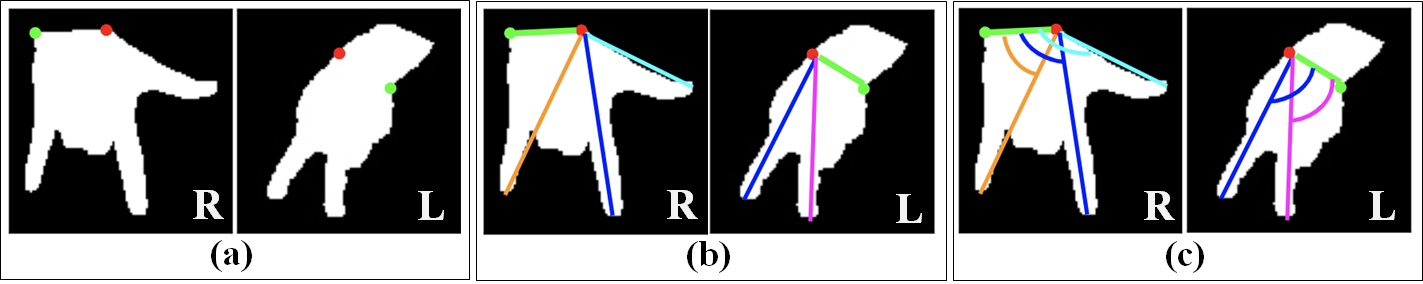,width=12cm}}
  \caption{Steps for hand keypoints misordering correction: (a) Depending on the left-right-hand prediction, the wrist point closest to the thumb is selected as `origin' (the red dot).  (b) Wrist-line and finger lines are drawn by connecting origin with other wrist point and fingertip respectively. (c) Based on hand geometry, the thumb-line creates the biggest angle and the little-finger-line creates the smallest angle when joined with the wrist-line.}
  \label{fig:fingertips misorder}
\end{figure}

\textbf{Fingertips localization:} Once the model is trained, the network starts predicting keypoint locations as the output of the third branch of the network. However, multiple pixels are predicted as `non-zero' in each channel of the output keypoint maps. One possible approach, as in \cite{yolse}, for localizing a fingertip in each channel is removal of all `non-zero' pixels with a value less than a threshold $p$, and selection of the pixel with the highest non-zero value (if any remaining pixel after removing pixels with value $< p$ ). However, finding an ideal threshold can be difficult. Moreover, often valid fingertips are rejected because of predicted pixel values lower than the threshold resulting lower prediction accuracy.

We adopted a different approach to resolve this problem. We defined a filtering function that takes advantage of the predictions in the gesture classification branch and handedness detection branch. The core idea is, given a predicted gesture, we can easily determine which channels should predict valid fingertips and which are not. For the channels with valid fingertips, we select the pixels with the highest prediction values. The second part of the function solves the finger index misorder problem. For example, a channel dedicated for index finger may predict the fingertip location of the middle finger and vise versa. This may happen because the two fingers are closely adjacent to each other and a small prediction error can swap their corresponding channels. To resolve this problem we utilize the predicted handedness information. Depending on whether the prediction is for a left-hand or a right-hand image, our function locates the wrist point closer to the thumb and connect it with the other wrist point (wrist-line), and all fingertip points (finger-lines). Now each finger-line creates an angle with the wrist-line. Because of the hand geometry, highest to lowest angels are produced by thumb-line, index-line, middle-line, ring-line, and little-finger-line respectively. We use this constraint to correct the misordered fingertips. The concept is highlighted in Figure \ref{fig:fingertips misorder}.  

\textbf{Wrist points localization:} Wrist points are filtered using a different approach. The location corresponding to the highest pixel value in the sixth channel is assigned to the first wrist point. The location of the second-highest pixel value, which is $> d_{th}$ pixel distance away from the first wrist point, is assigned to the second wrist point. The distance threshold is used to impose the condition that the two wrist points should not be detected in close proximity. We empirically determined that $d_{th} = 5 pixel$ produces very good results.

\subsection{Loss function}
The total loss is defined as: $L = \alpha L_{keypoints}+\beta L_{gesture}+\gamma L_{handedness}$.
The loss function has three parts corresponding to the three branches of the network. $L_{gesture}$ corresponding to branch 1 that estimates the mean gesture classification error. A categorical cross-entropy function is used to compute this. $L_{handedness}$ is the loss corresponding to the left-right hand detection branch, and a binary cross-entropy function is used for this. Finally, $L_{keypoints}$ loss is computed after the fingertips and wrist points localization branch. A mean squared error is used to compute this. The parameters $\alpha$, $\beta$, and $\gamma$ are empirically estimated using a grid search method. We used $\alpha=0.77$ $\beta=0.15$ and $\gamma=0.08$.

\begin{table}[t] 
\begin{center}
\caption{Ablation study for the gesture, hand keypoints, and handedness detection tasks.} \label{table:Ablation}
\begin{tabular}{|c|c|c|c|c|c|c|c|c|c|c|c|c|}
  \hline
  &  \multicolumn{3}{c|}{Gesture} & \multicolumn{3}{c|}{Fingertips} & \multicolumn{3}{c|}{Wrists} & \multicolumn{3}{c|}{Handedness} \\ 
  \hline
  & Recall & Prec. & Acc. & Recall & Prec. & Acc. & Recall & Prec. & Acc. & Recall & Prec. & Acc. \\ 
  \hline
  each branch & 98.23 & 98.27 & 98.23 & 83.98 & 96.49 & 91.32 & 98.36 & 92.74 & 91.33 & 98.71 & 98.77 & 98.72  \\
  all branch & 98.36 & 98.37 & 98.34 & 98.78 & 98.51 & 98.96 & 98.35 & 92.78 & 91.36 & 99.71 & 99.72 & 99.72 \\
  %w/ Aug. & 98.56 & 98.50 & 99.01 & 99.02 & 98.80 \\
  \hline
\end{tabular}
\end{center}
\end{table}

\section{Experiment}
Our training dataset consists of frames from 20 users, and the test dataset is formed by frames from the last 4 users. All the results reported in this section are on the test dataset. Additionally, We evaluated the performance of our model on four online datasets \cite{ExternalData-01,ExternalData-02,ExternalData-03,Mantecn2019}. All the experiments are performed on a desktop with Windows 10 operating system, an AMD RYZEN 9 3900X CPU, and a GTX 1080ti graphics card. For training our model, a batch size of 32 and a training epoch of 100 is used. Optimal model parameters are selected based on the cross-validation results on the training dataset.

\textbf{Ablation study:} We computed the accuracy when separate models are trained for gesture classification, keypoints detection, and handedness prediction. Table \ref{table:Ablation} summarized the results and compared them with a case when all branches are trained together. As it can be seen, the results are comparable for single branch and multi-task learning except for the fingertips detection. Fingertips detection heavily benefited from the information feedback of the other two branches.

\textbf{Gesture classification:} We computed the gesture classification performance of our method on our dataset and 3 external datasets \cite{ExternalData-01,ExternalData-02,ExternalData-03}. To compute the accuracy on the external datasets, we trained the model on the training dataset and generated the test results on those datasets. Since all datasets do not have the same 10 gestures as in our training dataset, we could only compute the accuracy for gestures that are common in the training dataset and external datasets. Some of the datasets are challenging because of the noisy hand segmentation data. Finally, we compared our results with well-known classification networks' accuracy. All the results are reported in Table \ref{table:gestureClassification}. Our network provides the best accuracy on the external datasets and near best accuracy on the self-collected dataset. This suggests the multi-task network can learn the tasks in a generalized manner.

\begin{table}[t]
\begin{center}
\caption{Hand gesture classification performance comparisons.} \label{table:gestureClassification}
\begin{tabular}{|c|c|c|c|c|c|c|c|c|c|c|c|c|}
  \hline
  Dataset &  Method & $G_1$ & $G_2$ & $G_3$ & $G_4$ & $G_5$ & $G_6$ & $G_7$ & $G_8$ & $G_9$ & $G_10$ & Avg. acc.  \\ 
  \hline

  \multirow{4}{*}{$D_1$ \cite{ExternalData-01}} & VGG16 & 100 & 75.7 & N/A & 97.7 & 84.3 & 100 & N/A & 99.3 & 100 & N/A & 93.86  \\
                        & MobileNet & 100 & 97.7 & N/A & 38.0 & 99.7 & 100 & N/A & 95.0 & 100 & N/A & 90.05  \\
                        & InceptionNet & 100 & 58.0 & N/A & 62.7 & 74.0 & 100 & N/A & 66.7 & 100 & N/A & 80.19  \\
                        & \textbf{Ours} & 100 & 99.7 & N/A & 94.7 & 96.3 & 100 & N/A & 100 & 100 & N/A & \textbf{98.67}  \\
  \hline                      
    \multirow{4}{*}{$D_2$ \cite{ExternalData-02}} & VGG16 & 82.8 & N/A & 33.8 & 41.0 & 83.8 & N/A & N/A & 84.3 & N/A & N/A & 64.76  \\
                        & MobileNet & 89.4 & N/A & 21.6 & 38.0 & 81.4 & N/A & N/A & 100 & N/A & N/A & 65.46  \\
                        & InceptionNet & 88.8 & N/A & 44.6 & 40.4 & 96.0 & N/A & N/A & 98.7 & N/A & N/A & 73.25  \\
                        & \textbf{Ours} & 88.4 & N/A & 60.0 & 71.0 & 97.4 & N/A & N/A & 93.6 & N/A & N/A & \textbf{81.87}  \\
  \hline                      
  \multirow{4}{*}{$D_3$ \cite{ExternalData-03}} & VGG16 & 65.5 & 43.6 & N/A & 84.7 & 82.9 & N/A & N/A & N/A & 35.0 & N/A & 63.90  \\
                        & MobileNet & 34.1 & 66.7 & N/A & 84.6 & 88.6 & N/A & N/A & N/A & 42.1 & N/A & 63.20  \\
                        & InceptionNet & 53.7 & 77.1 & N/A & 90.4 & 100 & N/A & N/A & N/A & 40.2 & N/A & 72.99  \\
                        & \textbf{Ours} & 72.5 & 98.0 & N/A & 99.0 & 93.7 & N/A & N/A & N/A & 78.5 & N/A & \textbf{88.32}  \\
 \hline                        
 \multirow{4}{*}{Self} & VGG16 & 94.9 & 97.5 & 97.6 & 85.9 & 99.0 & 99.4 & 95.2 & 95.9 & 97.3 & 88.7 & 95.17  \\
                        & MobileNet & 95.4 & 99.6 & 99.9 & 99.0 & 100 & 99.1 & 97.9 & 99.2 & 97.1 & 98.0 & \textbf{98.56}  \\
                        & InceptionNet & 99.0 & 99.1 & 96.9 & 95.7 & 100 & 100 & 94.0 & 98.1 & 95.4 & 98.6 & 97.68  \\
                        & \textbf{Ours} & 99.8 & 99.8 & 94.0 & 100 & 99.7 & 100 & 97.1 & 99.7 & 98.1 & 95.3 & 98.34  \\                         
  
  \hline
\end{tabular}
\end{center}
\end{table}

\textbf{Hand Keypoints detection:} We compare the hand keypoints localization performance of YOLSE \cite{yolse}, Unified Learning Approach (ULA) \cite{Alam2021}, and our method on an external dataset (D4) \cite{Mantecn2019} and self-collected dataset (Table \ref{table:HandKeypoints}). As it can be seen our results on both datasets for fingertips and wrist points localization are far superior compared to the state-of-the-art methods. For comparison, the forward pass time and the total number of network parameters for ULA are 18ms and 20.5 million, for YOLSE are 34ms and 2.86 million, and ours are 27ms and 6.19 million.    

\textbf{Handedness detection:} We compare handedness detection results with two well-known classification networks such as VGG16 \cite{vgg} and MobileNet \cite{mobilenet} (Table \ref{table:Handedness}). Our network performed almost as well as MobileNet. Noticeably, our network is much more lightweight compared to the other two networks. When VGG16 and MobileNet have 14.72 million and 3.23 million parameters respectively for a single branch, our network only has 6.19 million parameters for all three branches. Even though MobileNet produces accuracy which is slightly above our reported accuracy, it does not have any multitasking capability. Our results are significant in the sense that we can simultaneously produce very high accuracy for all the tasks.

\begin{table}[t]
\begin{center}
\caption{Keypoints detection results.} \label{table:HandKeypoints}
\begin{tabular}{|c|c|c|c|c|c|c|c|}
  \hline
Dataset & Methods & \multicolumn{3}{c|}{Fingertips} & \multicolumn{3}{c|}{Wrists} \\
\hline
& & Recall & Prec. & Acc. & Recall & Prec. & Acc. \\
\hline
\multirow{3}{*}{$D_4$ \cite{Mantecn2019}} & YOLSE & 66.58 & 73.84 & 70.03 & 88.95 & 90.82 & 81.61 \\
& ULA & 37.44 & 89.36 & 57.34 & 45.86 & 88.19 & 66.36 \\
& Ours & 84.59 & 88.91  & 86.71 & 96.55 & 86.86 & 84.24\\

\multirow{3}{*}{Self} & YOLSE & 74.70 & 85.11 & 81.87 & 94.16 & 89.31 & 84.62 \\
& ULA & 29.93 & 98.73 & 65.15 & 34.02 & 98.95 & 70.42 \\
& Ours & 98.51 & 98.78  & 98.96 & 98.35 & 92.78 & 91.36\\

\hline
\end{tabular}
\end{center}
\end{table}

\begin{table}[t]
\begin{center}
\caption{Handedness detection results.} \label{table:Handedness}
\begin{tabular}{|c|c|c|c|}
  \hline
  Methods & Recall & Prec. & Acc. \\
 \hline
 VGG16 & 96.45 & 96.73 & 96.50 \\
 MobileNet & 99.82 & 99.83 & 99.83 \\
 Ours & 99.71 & 99.72 & 99.72  \\
\hline
\end{tabular}
\end{center}
\end{table}

\section{Discussion and Conclusion }
In this paper, we introduced a multi-task network that simultaneously learns to predict hand gestures, hand keypoints, and handedness from thermal image inputs. Also, we collected a dataset of 24 users performing the gestures. In our experimental validation, we showed the effectiveness of the network as it learns to perform all the tasks with very high accuracy. We also showed that the network is able to learn the generalized concepts, and as a result, the network performs well on external datasets where other well-known networks fail.   

This work shows promise, especially in the current technology landscape where there are interests to interact with intelligent devices in a natural way such as using gestures. Moreover, our study is based on an alternative data modality that can be combined with color image data to build a better and more robust system. Our future research directions will explore those possibilities. To the best of our knowledge, this is the first attempt to simultaneously learn the three proposed tasks using a single DL network with thermal images. In the future, we will explore combining different modalities of data in a single DL network pipeline for better performance.    

Muti-task deep learning is advantageous because it allows simultaneous learning of multiple correlated tasks. Therefore, it serves as a method of regularisation because it encourages learning only the features relevant for all tasks in the shared part of the network. This helps generalized learning. Our network shows this trait as we showed that the network performed way better than the other models on external datasets. Muti-task learning can also be used as learning some related intermediate tasks, and using the knowledge of intermediate learning to boost the performance of the final tasks \cite{Xu2018}. One of our future research directions is adapting this idea to improve the performance of the three tasks we addressed in this work.      

\bibliographystyle{splncs03}
\bibliography{ICDEC_ref}

\end{document}